\documentclass[sigconf]{acmart}

\usepackage{xcolor}
\usepackage{xspace}
\usepackage{graphicx}
\usepackage{multirow} 
\usepackage{pifont}
\usepackage{makecell}
\usepackage{bm}

\copyrightyear{2025}
\acmYear{2025}
\setcopyright{acmlicensed}\acmConference[MM '25]{Proceedings of the 33rd ACM International Conference on Multimedia}{October 27--31, 2025}{Dublin, Ireland}
\acmBooktitle{Proceedings of the 33rd ACM International Conference on Multimedia (MM '25), October 27--31, 2025, Dublin, Ireland}
\acmDOI{10.1145/3746027.3755494}
\acmISBN{979-8-4007-2035-2/2025/10}

\begin{document}
\newcommand{\cmark}{\ding{51}\xspace}
\newcommand{\cmarkg}{\textcolor{lightgray}{\ding{51}}\xspace}
\newcommand{\xmark}{\ding{55}\xspace}
\newcommand{\xmarkg}{\textcolor{lightgray}{\ding{55}}\xspace}
\newcommand{\omark}{\ding{70}\xspace}

\newcommand\our{\textsc{\mbox{FELLE}}}

\title{FELLE: Autoregressive Speech Synthesis with Token-Wise Coarse-to-Fine Flow Matching}

\author{Hui Wang}
\orcid{0009-0003-8057-4644}
\authornote{Work done during an internship at Microsoft.
}
\affiliation{%
  \institution{College of Computer Science, Nankai University}
  \city{Tianjin}
  \country{China}
}

\author{Shujie Liu}
\orcid{0009-0008-2599-6752}
\authornote{Corresponding authors.}
\affiliation{%
  \institution{Microsoft Corporation}
  \city{Hong Kong}
  \country{China}
}

\author{Lingwei Meng}
\orcid{0000-0003-1028-6017}
\affiliation{%
  \institution{Microsoft Corporation}
  \city{Beijing}
  \country{China}
}

\author{Jinyu Li}
\authornotemark[2] 
\orcid{0000-0002-1089-9748}
\affiliation{%
  \institution{Microsoft Corporation}
  \city{Redmond}
  \country{United States}
}

\author{Yifan Yang}
\orcid{0009-0003-0588-1812}
\affiliation{%
  \institution{Microsoft Corporation}
  \city{Beijing}
  \country{China}
}

\author{Shiwan Zhao}
\orcid{0000-0001-5068-025X}
\affiliation{%
  \institution{College of Computer Science, Nankai University}
  \city{Tianjin}
  \country{China}
}

\author{Haiyang Sun}
\orcid{0009-0004-3485-3869}
\affiliation{%
  \institution{Microsoft Corporation}
  \city{Beijing}
  \country{China}
}

\author{Yanqing Liu}
\orcid{0000-0002-4150-0680}
\affiliation{%
  \institution{Microsoft Corporation}
  \city{Beijing}
  \country{China}
}

\author{Haoqin Sun}
\orcid{0000-0002-8554-8969}
\affiliation{%
  \institution{College of Computer Science, Nankai University}
  \city{Tianjin}
  \country{China}
}

\author{Jiaming Zhou}
\orcid{0009-0002-4819-4572}
\affiliation{%
  \institution{College of Computer Science, Nankai University}
  \city{Tianjin}
  \country{China}
}

\author{Yan Lu}
\orcid{0000-0001-5383-6424}
\affiliation{%
  \institution{Microsoft Corporation}
  \city{Beijing}
  \country{China}
}

\author{Yong Qin}
\authornotemark[2] 
\orcid{0009-0000-2748-3020}
\affiliation{%
  \institution{College of Computer Science, Nankai University}
  \city{Tianjin}
  \country{China}
}

\renewcommand{\shortauthors}{Hui Wang et al.}

\begin{abstract}
  To advance continuous token modeling and temporal-coherence enforcement, we propose \our{}, an autoregressive model that integrates language modeling with token-wise flow matching. By leveraging the autoregressive nature of language models and the generative efficacy of flow matching, \our{} effectively predicts continuous-valued tokens (mel-spectrograms). For each continuous-valued token, \our{} modifies the general prior distribution in flow matching by incorporating information from the previous step, improving coherence and stability. Furthermore, to enhance synthesis quality, \our{} introduces a \textit{coarse-to-fine} flow-matching mechanism, generating continuous-valued tokens hierarchically, conditioned on the language model’s output. Experimental results demonstrate the potential of incorporating flow-matching techniques in autoregressive mel-spectrogram modeling, leading to significant improvements in TTS generation quality, as shown in \url{https://aka.ms/felle}.
\end{abstract}



\maketitle

\begin{figure}[t]
  \includegraphics[width=\columnwidth]{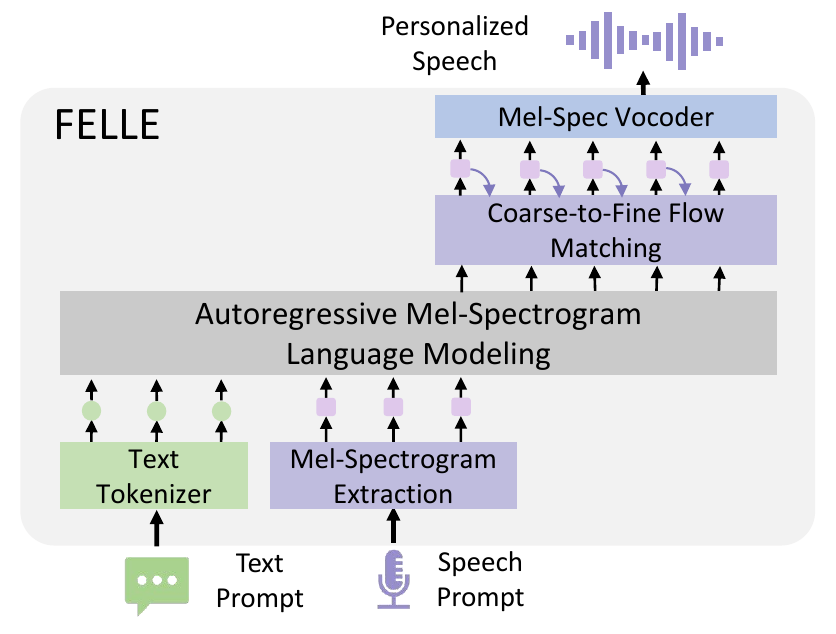}
  \caption{\our{} is an autoregressive mel-spectrogram model that generates personalized speech from text and acoustic prompts. It uses the previous mel-spectrogram as a prior and refines features with a coarse-to-fine flow-matching module guided by the language model.}
  \Description{Diagram showing the architecture of \our{}, an autoregressive model that generates personalized speech from text and acoustic prompts. The model predicts mel-spectrograms by conditioning on both previous outputs and language model features, using a coarse-to-fine flow-matching module for refinement.}
  \vspace{-10pt}
  \label{fig:overview}
\end{figure}

\section{Introduction}
The remarkable success of large language models (LLMs) \cite{brown2020language, achiam2023gpt, team2024gemini} has prompted a paradigm shift in speech synthesis, redefining it as a language modeling task. This shift has driven notable progress in zero-shot speech synthesis \citep{wang2023valle, chen2024valle2}. Consistent with the standard LLM training methodology, researchers have naturally adopted discrete-valued tokens as the foundational modeling units. However, unlike textual data, which is inherently discrete, speech signals require complex quantization techniques to transform continuous waveforms into discrete-valued tokens. These essential quantization processes impose fundamental constraints compared to continuous representations, particularly in terms of fidelity preservation and training complexity \citep{puvvada2024discrete, meng2024autoregressive}. Consequently, discrete token-based text-to-speech (TTS) systems often face challenges such as intricate modeling workflows and reduced output quality. In response to these limitations, recent research has increasingly explored autoregressive (AR) modeling frameworks that leverage continuous representations \citep{meng2024autoregressive, turetzky2024continuous, zhu2024autoregressive,sled}, showing notable improvements in model performance and simplifying training processes.

However, modeling continuous representations introduces its own set of challenges. Due to the rich information contained in continuous representations, modeling them demands more advanced capabilities from models. Conventional regression-based loss functions used in MELLE~\citep{meng2024autoregressive}, including mean absolute error (MAE) and mean squared error (MSE), adopt oversimplified distributional assumptions. These assumptions may not fully capture the multimodal structures and complex features of the distribution, leading to blurred, oversimplified, or averaged predictions \citep{vasquez2019melnet,ren-etal-2022-revisiting}. Similarly, KALL-E relies on WaveVAE-derived distributions, but the restrictive Gaussian prior assumption in variational autoencoder (VAE)~\citep{kingma2013auto} limits their ability to model complex speech patterns, leading to low-diversity and blurry samples \citep{tomczak2018vae, bredell2023explicitly}.

A further limitation of existing approaches lies in the inadequate modeling of temporal dependencies. Current methodologies primarily use autoregressive architecture to implicitly capture temporal dependencies, yet they lack explicit mechanisms to model temporal relationships. This structural characteristic may limit their effectiveness in handling complex temporal dependencies~\citep{han2024valler}. For instance, SALAD~\citep{turetzky2024continuous}, which is based on diffusion processes, denoises tokens independently without explicit temporal modeling. MELLE~\citep{meng2024autoregressive} applies a flux loss focused solely on increasing frame-level variability, oversimplifying the modeling of temporal relationships. Notably, continuous-valued tokens like mel-spectrograms inherently exhibit strong correlations across temporal and frequency dimensions \citep{ren-etal-2022-revisiting}. Insufficient consideration of these correlations could compromise the model's ability to preserve speech's sequential characteristics, potentially affecting output naturalness and requiring additional computational resources.

In this work, we introduce \our, an autoregressive speech synthesis framework that utilizes token-wise coarse-to-fine flow matching for continuous-valued token modeling. Unlike regression-based or VAE approaches (commonly used in other methods) constrained with preset distribution assumptions, flow matching~\citep{lipman2022flow} enables flexible density estimation without restrictive prior assumptions, thereby preserving the multimodal characteristics of speech. Meanwhile, by integrating the autoregressive properties of language models with flow-matching techniques, we develop a temporal modeling mechanism that dynamically adjusts the prior distribution of each frame through the integration of preceding contextual information. This architecture effectively preserves temporal dependencies and ensures spectral continuity. Moreover, we propose a coarse-to-fine flow-matching (C2F-FM) module to improve generation quality by capturing inter-frequency correlations. It synthesizes mel-spectrogram features in multiple stages, inspired by the effectiveness of coarse-to-fine methods in discrete token modeling \citep{borsos2023audiolm, defossez2024moshi}, which capture structural dependencies in sequential tasks. Evaluations on the LibriSpeech corpus~\citep{panayotov2015librispeech} demonstrate the framework's competitiveness: compared to MELLE, our method achieves comparable Word Error Rates (WER) while delivering superior similarity scores in modeling complex mel-spectrogram patterns. Our contributions can be summarized as:
\begin{itemize}
    \item We propose an AR speech synthesis framework leveraging token-wise flow matching for continuous speech modeling, eliminating restrictive distribution assumptions while preserving speech signals' multimodal characteristics.
    \item We design a dynamic prior mechanism that modifies the vanilla prior distribution in flow matching by incorporating information from the previous step, improving coherence and stability.
    \item We introduce a coarse-to-fine flow matching architecture that explicitly captures inter-frequency correlations through multi-stage spectral refinement, achieving significant improvements in mel-spectrogram generation.
\end{itemize}

\section{Related Work}

Zero-shot TTS are commonly categorized into autoregressive and non-autoregressive paradigms based on their output generation mechanisms. Autoregressive systems typically rely on language model architectures \citep{wang2023valle, kharitonov2023speak,yang2024interleaved}, whereas non-autoregressive implementations commonly employ diffusion models and analogous methodologies \citep{junaturalspeech, chen2024f5,wang2025maskgct}. The subsequent discussion focuses on research efforts investigating diverse representations under the framework of autoregressive language modeling architectures.

\subsection{Discrete-Valued Token-Based TTS}
TTS systems based on discrete representations \citep{wang2023valle, lajszczak2024base, song2024ellav,du2024cosyvoice, du2024cosyvoice2} utilize tokenized acoustic units derived from unsupervised or semi-supervised learning frameworks. These discrete tokens serve as compact representations of speech, capturing phonetic and prosodic attributes while reducing redundancy in data storage and computation. VALL-E \citep{wang2023valle} is a neural codec language model for text-to-speech synthesis that firstly redefines TTS as a conditional language modeling task, enabling high-quality, personalized speech generation from just a 3-second acoustic prompt, significantly advancing naturalness and speaker similarity. Recent studies further enhance VALL-E’s capabilities across multilingual generalization \citep{zhang2023vallex}, decoding efficiency \citep{chen2024valle2}, and robustness \citep{song2024ellav,xin2024ralle,han2024valler}, collectively advancing zero-shot TTS in scalability, quality, and linguistic flexibility. In contrast to the unified language modeling approach of VALL-E and its variants, CosyVoice \citep{du2024cosyvoice} leverages an LLM for text-to-token conversion followed by a conditional flow-matching model for token-to-spectrogram synthesis, enhancing zero-shot voice cloning through end-to-end supervised speech token learning.

\subsection{Continuous-Valued Token-Based TTS}
Recent advances in continuous representation-based TTS systems eliminate the need for cumbersome codec training while achieving promising performance. Notably, MELLE \citep{meng2024autoregressive} proposes a single-pass language model architecture leveraging rich continuous acoustic representations, enabling precise control over prosodic features including pitch, rhythm, and timbre for high-fidelity speech synthesis. SALAD \citep{turetzky2024continuous} is a zero-shot text-to-speech system that employs a per-token latent diffusion model on continuous representations, enabling variable-length audio generation through semantic tokens for contextual guidance and stopping control. While this method achieves superior intelligibility scores, it may face challenges related to time costs. Alternatively, KALL-E \citep{zhu2024autoregressive} adopts an autoregressive approach with WaveVAE to directly model speech distributions, bypassing both VAE and diffusion paradigms, demonstrating enhanced performance through probabilistic waveform prediction.

\section{Preliminary}
\subsection{Background}
\paragraph{Flow Matching} 
Flow matching~\citep{lipman2022flow} is a technique for learning a transformation that maps a prior distribution \( p_0 \) to a target distribution \( q(x) \). The core idea of flow matching is to define a flow \( \phi_t(x) \) that evolves over time, transforming the prior distribution \( p_0 \) into the target distribution \( q(x) \). This flow \( \phi_t(x) \) is governed by a vector field \( v_t(x) \) and satisfies the following ordinary differential equation:
\(\frac{d}{dt} \phi_t(x) = v_t(\phi_t(x)),\quad \phi_0(x) = x.\)
Here, \( \phi_0(x) = x \) indicates that at time \( t = 0 \), the flow \( \phi_t(x) \) is an identity mapping.

While flow matching provides a principled framework for learning such transformations, it can be computationally expensive due to the difficulty of directly accessing the true vector field \( u_t(x) \) and the target distribution \( q(x) \). To address this, Conditional Flow Matching (CFM) is introduced. In CFM, the flow and the vector field are conditioned on the data \( x_1 \), making the optimization process more efficient. The objective of CFM is to minimize the discrepancy between the conditional true vector field \( u_t \) and the learned conditional vector field \( v_t(x; \theta) \). This discrepancy is measured by the following loss function:
$
L_{\text{CFM}} = \mathbb{E}_{t, x_1, x} \left\| u_t - v_t(x; \theta) \right\|^2,
$
where time \( t \) is uniformly sampled from \( \mathcal{U}[0,1] \), data points \( x_1 \) are drawn from the target distribution \( q(x_1) \), samples \( x \) are generated through the conditional probability path \( p_t(x|x_1) \), and the conditional vector field \( u_t \equiv u_t(x|x_1) \). 


\subsection{Problem Formulation}
Following MELLE's autoregressive language modeling framework for mel-spectrogram prediction, we reformulate zero-shot TTS through a hierarchical flow-matching mechanism at each prediction step. Each mel-spectrogram frame \(\bm{x}^i \in \mathbb{R}^D\) (where \(D\) denotes the mel-band dimension) is treated as a continuous token, generated sequentially through an autoregressive process. Given an input text sequence \(\bm{y} = [y^0, \ldots, y^{N-1}]\), speech prompt \(\bm{\widehat{x}}\), and previously generated tokens \(\bm{x}^{<i} = [\bm{x}^0, \ldots, \bm{x}^{i-1}]\), the model predicts the current token \(\bm{x}^i\) by integrating language model guidance into the flow-matching paradigm. The joint distribution is decomposed autoregressively as:
\begin{align}
    p(\bm{X} \! \mid\!\bm{y})\!  
   &= \prod_{i=0}^{L-1} p(\bm{x}^i \mid \bm{x}^{<i}, \bm{y}, \bm{\widehat{x}})  \\
   &=\! \prod_{i=0}^{L-1} p_{\theta_\text{FM}}(\bm{x}^i \mid \bm{z}^i), \bm{z}^i\!=\!f_{\theta_\text{LM}}(\bm{x}^{<i}, \bm{y},\bm{\widehat{x}} \notag) .
\end{align}
\(\bm{X} = [\bm{x}^0, \ldots, \bm{x}^{L-1}] \in \mathbb{R}^{L \times D}\) denotes full mel-spectrogram sequence, \(L\) represents the total number of mel-spectrogram frames. The language model \(f_{\theta_\text{LM}}(\cdot)\) generates hidden state \(\bm{z}^i\) that captures both linguistic content and acoustic context, while \(p_{\theta_\text{FM}}(\cdot \mid \bm{z}^i)\) denotes the flow-matching module that transforms prior distributions into target distributions conditioned on \(\bm{z}^i\).

\section{\our{} Architecture}

The proposed framework combines an autoregressive language model with a flow-matching mechanism, which facilitates the progressive generation of high-fidelity speech. As shown in Figure~\ref{fig:overview}, the autoregressive model $f_{\theta_\text{LM}}$ extracts features from the text prompt \( \bm{y} \) and speech prompt  \( \bm{\widehat{x}} \), generating latent representations \(\bm{z}^i\) (where \(i\) denotes the generation step) that serve as conditional inputs for the flow-matching mechanism. The flow-matching mechanism applies a coarse-to-fine strategy to generate high-quality mel-spectrogram frames $\bm{x}^i$. The main components are described in detail below.

\begin{figure*}[t]
    \centering
    \includegraphics[width=0.8\linewidth]{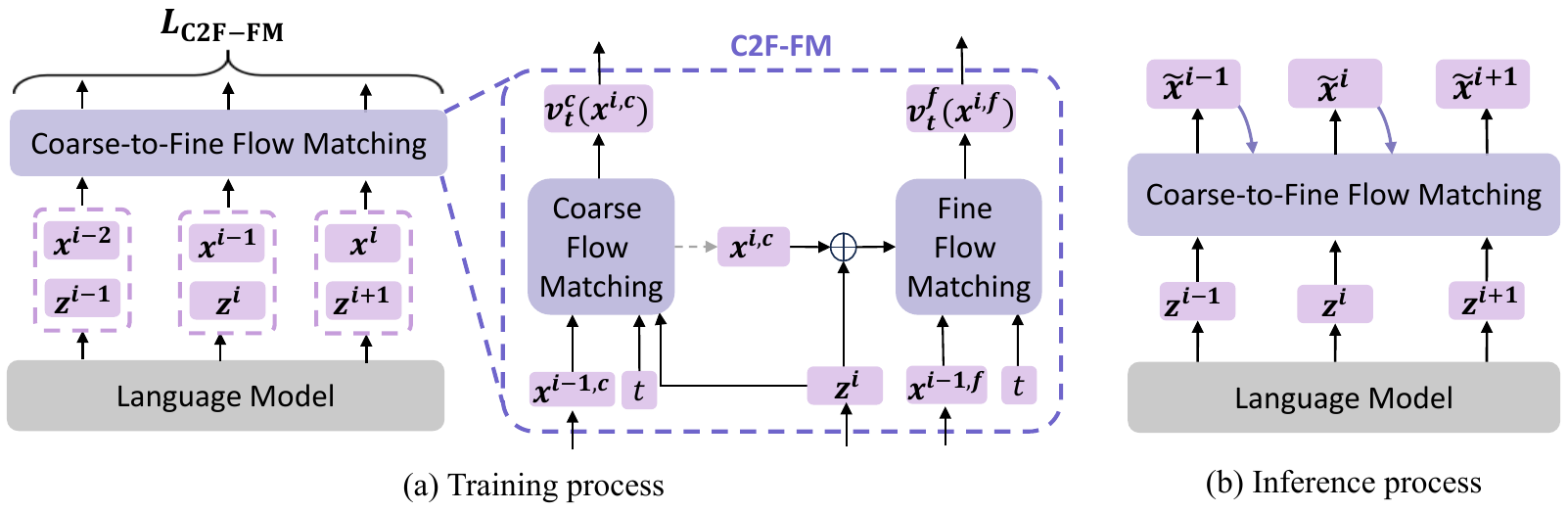}
    \caption{The \textit{coarse-to-fine flow-matching} module of \our. (a) The training process along with the detailed data flow within the coarse-to-fine module. The gray dashed lines merely indicate the relationships between components in the model structure and are not activated during training. (b) The inference process.}
    \Description{Two-part diagram of the coarse-to-fine flow-matching module in \our. Part (a) shows the training process, illustrating the internal data flow through the coarse and fine stages. Gray dashed lines represent inactive connections used only for model structure reference. Part (b) depicts the inference process with active data paths.}
    \vspace{-5pt}
    \label{fig:Coarse-to-Fine}
\end{figure*}

\subsection{Autoregressive Language Model}
The language model, designed as a Transformer decoder, generates acoustic features autoregressively by utilizing both text sequences and mel-spectrogram prompts. In the initial step, the text tokens are embedded, while a pre-net maps the mel-spectrogram into the dimensional space of the LM. By processing the combined text \( \bm{y} \), speech prompt \( \bm{\widehat{x}} \), and acoustic embeddings \( \bm{x}^{<i} \), the language model \( f_{\theta_\text{LM}} \) processes multi-head attention and feed-forward layers to capture the intricate relationship between linguistic and acoustic information. The output at each time step subsequently serves as a conditioning input for the coarse-to-fine flow-matching module to synthesize the next-frame acoustic features.

\subsection{Coarse-to-Fine Flow Matching}
For high-quality mel-spectrogram generation, we introduce a coarse-to-fine flow-matching approach. As illustrated in Figure~\ref{fig:Coarse-to-Fine}, the method generates each mel-spectrogram frame based on its preceding frame, maintaining temporal consistency throughout the sequence. The generation process is divided into two phases: a coarse generation phase followed by a fine refinement phase. A detailed introduction will be given below.

\paragraph{Prior Distribution}  
Flow-matching-based methods in speech synthesis commonly adopt a simple prior distribution \citep{le2024voicebox, mehta2024matcha}, as prior knowledge is often challenging to define precisely \citep{chen2024f5}. However, utilizing a prior distribution that closely aligns with the target distribution can significantly enhance computational efficiency and synthesis quality \citep{zhang2024speechgpt}. Given the autoregressive nature of token generation and the sequential structure of speech, \our{} employs the preceding token as an informative prior to guide the flow matching process for generating the current token. Specifically, the prior distribution \( p_0 \) for the initial state \( x_0^i \) of the current frame \( x^i \) is derived from the mel-spectrogram of the previous frame \( x^{i-1} \):
\begin{align}
\label{eq:prior}
p_0(x_0^i | x^{i-1}) = \mathcal{N}(x_0^i | x^{i-1}, \sigma^2 I),
\end{align}
where \( \sigma^2 I \) represents the covariance matrix of the Gaussian noise. For \( i = 0 \), where no prior frame exists, the initial state is drawn from a standard Gaussian distribution.

\paragraph{Coarse-to-Fine Generation}  
Our method combines autoregressive language modeling with hierarchical flow matching. Each step \( i \) follows a two-stage process, as illustrated in Figure~\ref{fig:Coarse-to-Fine}(a): a coarse flow-matching phase that produces an initial low-resolution mel-spectrogram representation, followed by a fine flow-matching phase that enhances the output by incorporating both the coarse representation and language model outputs.

The coarse generation stage is designed to produce the low-resolution component \( x^{i,c} \) of the \( i \)-th frame through a downsampling operation \( x^{i,c} = \mathrm{Downsample}(x^i) \). In this framework, the coarse flow-matching model predicts a vector field \( v_t^c(x^{i,c}, z^i; \theta_{\text{FM}}^c) \) by conditioning on linguistic features \( z^i \) extracted from the language model. 

In the fine stage, the model further refines this approximation by recovering fine-grained details \( x^{i,f} \), represented as the residual between the original frame \( x^i \) and the upsampled coarse component \( \text{\textit{Upsample}}(x^{i,c}) \). A secondary flow-matching model predicts the vector field  \( v_t^f(x^{i,f}, z^i, x^{i,c}; \theta_\text{FM}^f) \), governing this process by leveraging both the features \( z^i \) and the coarse component (with ground-truth coarse features $x^{i,c}$ during training and predicted values $\tilde{x}^{i,c}$ during inference) as conditional inputs. This hierarchical conditioning allows the fine model to focus on local details while preserving global coherence from the coarse stage.

For step \( i \), the training objective combines losses from both stages:
\begin{align}
&\mathcal{L}_{\text{C2F-FM}}\!=\! \mathbb{E}_{t, x_1^{i,c}, x^{i,c}} \big\| u_t^c - v_t^c(x^{i,c}, z^i; \theta_\text{FM}^c) \big\|^2 \notag \\
&+\mathbb{E}_{t, x_1^{i,f}, x^{i,f}} \big\| u_t^f \!- \!v_t^f(x^{i,f}, z^i, x^{i,c}; \theta_\text{FM}^f) \big\|^2,
\end{align}
where \( u_t^c \) and \( u_t^f \) represent the true conditional vector fields for the coarse and fine components, respectively, and \( t \sim \mathcal{U}[0,1] \). The initial states \( x_0^{i,c} \) and \( x_0^{i,f} \) are similarly initialized using the prior from Equation~\ref{eq:prior}, applying the corresponding sampling operations. By decoupling low-resolution structure learning from high-detail refinement, this coarse-to-fine approach generates high-fidelity mel-spectrograms while maintaining temporal consistency through autoregressive dependencies.

\paragraph{Classifier-Free Guidance}  
Classifier-free guidance (CFG) is a powerful technique to enhance the quality and controllability of generated outputs in flow matching and diffusion models \citep{ho2022classifier, nichol2021improved}. In \our, we implement CFG through joint training of coarse and fine flow matching models using both conditional and unconditional objectives. During training, we randomly mask the speech prompt with probability $p_{\text{drop}}$ for unconditional learning, which enables each model to learn dual vector fields. At inference, guided vector fields are computed by linear blending:  
\begin{align}
\hat{v}_t^\ast(x^\ast; \cdot) &= w v_t^\ast(x^\ast, c; \theta_\text{FM}^\ast) + (1-w) v_t^\ast(x^\ast, \bar{c}; \theta_\text{FM}^\ast),
\end{align}  
where $\ast \in \{c,f\}$ denotes the model stage, \(c\) represents the full conditions, \(\bar{c}\) indicates the reduced conditioning state where the speaker prompt is masked, and \(w\) represents the guidance scale.

\subsection{Training Objective}

In \our, we integrate the condition loss \( \mathcal{L}_\text{cond} \) in addition to coarse-to-fine loss \( \mathcal{L}_{\text{C2F-FM}} \).  \( \mathcal{L}_\text{cond} \) is a hybrid loss function that combines L1 and L2 norms, defined as \( \mathcal{L}_{\text{cond}} = \|z_i - x_i\|_1 + \|z_i - x_i\|_2^2 \), for step $i$ to regularize the conditional input for flow matching. Additionally, we introduce a stop prediction module to the autoregressive language model. This module, during each step of generation, transforms the hidden state output by the language model into the probability of a stop signal through a linear layer and calculates the Binary Cross-Entropy loss \( \mathcal{L}_\text{stop} \) for training. The model can automatically determine when to stop during the generation process without the need to preset length rules. The overall training objective is:
$\mathcal{L} = \mathcal{L}_{\text{C2F-FM}} + \lambda \mathcal{L}_{\text{cond}} + \alpha \mathcal{L}_{\text{stop}},$
where \( \lambda \) and \( \alpha \) control the respective contributions of \( \mathcal{L}_{\text{cond}} \) and \( \mathcal{L}_{\text{stop}} \).


\subsection{Inference}

As illustrated in Figure~\ref{fig:Coarse-to-Fine}(b), the inference process employs an autoregressive language model that progressively generates hidden representations based on textual and speaker prompts. At each step \(i\), the computed latent state \(z_i\) serves two key purposes. First, it provides conditional guidance for the coarse flow-matching module, facilitating the gradual transformation from the previous mel-spectrogram approximation \(\tilde{x}^{i-1,c}\) to the current coarse structural estimate \(\tilde{x}^{i,c}\). Following this coarse estimation phase, the integrated information of \(\tilde{x}^{i,c}\) and \(z_i\) drives the fine flow-matching module to produce the fined mel-spectrogram frame \(\tilde{x}^{i,f}\). The final output frame \(\tilde{x}^i\) emerges through the integration of these complementary coarse and refined predictions. Secondly, the latent state \(z_i\) processed by the stop prediction module to compute the stop probability, which is compared against a predefined threshold to decide whether to terminate the process. The iterative generation continues until the stop criterion is satisfied, after which a neural vocoder converts the mel-spectrogram into the final speech waveform.

\section{Experimental Setup}

\subsection{Dataset}

We employ the LibriSpeech dataset \citep{panayotov2015librispeech} for \our{} training. LibriSpeech consists of approximately 960 hours of speech data sourced from audiobooks available on the LibriVox platform. It features recordings from 1,251 speakers, showcasing a wide range of accents, intonations, and speaking styles. For textual representation, we utilize phoneme-based tokens. On the audio side, 16 kHz waveforms are processed to extract 80-dimensional log-magnitude mel-spectrograms through a short-time Fourier transform (STFT) and an 80-dimensional mel filter, covering a frequency range from 80 Hz to 7,600 Hz. The acoustic representation is finalized by applying a base-10 logarithm to the extracted features.

For zero-shot text-to-speech evaluation, we use the LibriSpeech test-clean set, ensuring that its speakers are entirely excluded from the training data. Following recent works \citep{han2024valler, meng2024autoregressive}, we select audio samples ranging from 4 to 10 seconds in duration for evaluation.

\subsection{Model Detail}
\paragraph{Model Configurations}
Our model consists of 12 Transformer blocks, designed in line with the architectures of VALL-E \citep{wang2023valle} and MELLE \citep{meng2024autoregressive} to ensure a fair comparison. Each block features 16 attention heads and a feed-forward layer with a dimensionality of 4,096. The decoder incorporates an embedding dimension of 1,024 and ReLU activation function. The input mel-spectrograms are transformed into the model's embedding space using a three-layer fully connected network. HiFi-GAN vocoder\footnote{\url{https://huggingface.co/mechanicalsea/speecht5-tts}} \citep{kong2020hifigan} is used for audio reconstruction.

The downsampling operation preserves the even-indexed mel-spectrogram frames as the coarse components. During upsampling, these components are expanded through zero-insertion at odd-indexed positions. Both coarse and fine flow-matching stages share identical backbone architectures comprising three residual blocks, each containing layer normalization, dual fully connected layers, and SiLU activation. The timestep embedding module combines sinusoidal positional encoding with two fully connected layers and SiLU activation. Key architectural differences emerge in the conditioning modules: the coarse stage uses single linear projections for language model outputs, and the fine stage incorporates additional layers to integrate auxiliary features like coarse-mel information.

\paragraph{Training and Inference Details}  The model is trained for 2 million iterations on 8 NVIDIA V100 GPUs. Loss coefficients are set to \(\beta = 0.1\) for the condition loss \(\mathcal{L}_{\text{cond}}\) and \(\sigma = 0.01\) for the stop prediction loss \(\mathcal{L}_{\text{stop}}\). The noise variance during training is configured at 0.1. CFG is implemented with dropout probability $p_{\text{drop}} = 0.1$. For unconditional setting, we apply a mask of random length between 3 and 10 seconds to speech prompt. During the inference process, we simultaneously process two types of input: one with complete speech prompts and another with masked speech prompts, used as classifier-free guidance. The results are then combined using a weighting factor of \(w = 1.6\) to produce the final output. The flow-matching framework performs 3 function evaluations using Euler’s method to iteratively generate mel-spectrograms.

\subsection{Evaluation Setting}
The performance of \our{} is evaluated under two distinct inference schemes to comprehensively assess its capabilities. 

\paragraph{Continuation:} Using the text transcription and the initial 3 seconds of the utterance as a prompt, the model is tasked with seamlessly synthesizing the continuation of the speech.
 
\paragraph{Cross-sentence:} Given a reference utterance and its transcription from the same speaker as the prompt, along with the text of the target utterance, the model is expected to synthesize the corresponding speech while preserving the speaker's characteristics.

\subsection{Evaluation Metric} 

\paragraph{Word Error Rate (WER)}: We use two automated speech recognition (ASR) models, the Conformer-Transducer model\footnote{\url{https://huggingface.co/nvidia/stt_en_conformer_transducer_xlarge}} \citep{gulati20conformer} and the HuBERT-Large ASR model\footnote{\url{https://huggingface.co/facebook/hubert-large-ls960-ft}} \citep{hsu2021hubert}, to evaluate the robustness and intelligibility of the synthesized speech. By contrasting the transcriptions produced by these models with the matching ground truth, the WER is determined. In particular, the WER scores derived from the Conformer-Transducer and HuBERT-Large systems are indicated by WER-C and WER-H, respectively.

\paragraph{Speaker Similarity (SIM)}: WavLM-TDNN\footnote{\url{https://github.com/microsoft/UniSpeech/tree/main/downstreams/speaker\_verification}} \citep{chen2022wavlm} is employed to extract speaker embeddings from the reference speech and the synthesized speech to assess the in-context learning capability of zero-shot TTS models. The cosine distance, which goes from -1 to 1, is used to measure how similar these embeddings are to one another. There are two assessment indicators taken into account: SIM-r measuring the similarity between synthesized speech and the reconstructed speech prompt, and SIM-o comparing synthesized speech with the original prompt.

\paragraph{Mean Opinion Score (MOS)}: MOS is a widely used metric that reflects the perceived quality of speech, typically rated by listeners on a scale from 1 (bad) to 5 (excellent). With the rapid development of automatic MOS prediction technology \citep{9746395, wang24s_interspeech, wang2023intermediate, liu2025musiceval}, it is now possible to evaluate quality accurately and effectively. We use the RAMP+ model\footnote{\url{https://github.com/NKU-HLT/RAMP\_MOS}} \citep{wang23r_interspeech,10933540} for speech quality assessment.


\begin{table}
\caption{The predicted MOS results.}
\vspace{-5pt}
\centering
\resizebox{0.9\linewidth}{!}{%
\begin{tabular}{c cc}
\toprule
\textbf{System} & \textbf{Continuation} & \textbf{Cross-Sentence} \\
\midrule
Ground Truth & $4.043_{\pm 0.32} $ & $4.043 _{\pm 0.32} $ \\
\midrule
VALL-E  & $1.828_{\pm 0.24}$ & $1.965_{\pm 0.27}$ \\
MELLE  & $\textbf{3.843}_{\pm 0.38}$ & $4.036_{\pm 0.25}$ \\
\midrule
\our & $3.836_{\pm 0.39}$ & $\textbf{4.157}_{\pm 0.19}$ \\
\bottomrule
\end{tabular}%
}

\label{tab:ramp_mos}
\end{table}

\begin{table*}


  \caption{The objective performance comparison of the \textit{continuation} and \textit{cross-sentence} zero-shot speech synthesis tasks, with WER-C (\%) and WER-H (\%) as WER metrics. *The reproduction results are quoted from \cite{han2024valler}. \textsuperscript{\dag}Results reported in \cite{meng2024autoregressive} are used. \textsuperscript{\ddag}\cite{meng2024autoregressive} provides results of MELLE across various setups, and we show the performance on the same dataset.
  }

  \label{tab:obj_1}

  \centering 
    \resizebox{0.8\textwidth}{!}
  {
  \begin{tabular}{l cccc cccc}

    \toprule
 \multirow{2}{*}{\textbf{System}} & \multicolumn{4}{c}{\textbf{Continuation}} & \multicolumn{4}{c}{\textbf{Cross-Sentence}} \\
 \cmidrule(r){2-5} \cmidrule(r){6-9}
      & WER-C  & WER-H  & SIM-r & SIM-o   & WER-C  & WER-H &  SIM-r & SIM-o   \\
  \midrule
   Ground Truth &  1.61  & 2.15 & - & 0.668  & 1.61  & 2.15 & - & 0.779\\
   Ground Truth (mel) & 1.64  & 2.24 & 0.622 & 0.617 & 1.64  & 2.24 & 0.747 & 0.732 \\
    \midrule
     VALL-E \citep{wang2023valle}  & -  &  3.8 & 0.508 & - & -  & 5.9 & 0.580 & - \\
    ELLA-V  \citep{song2024ellav} * & 2.10 & 2.91 & 0.340 & 0.303  & 7.15 & 8.90 & 0.331 & 0.307 \\
    RALL-E \citep{xin2024ralle} &-&-&-&-& 2.50  & \textbf{2.80} & - &0.49   \\
    CLaM-TTS \citep{kim2024clamtts} & -  & 2.36 & 0.513 & 0.477 & -  & 5.11 & 0.538 & 0.495 \\
    VALL-E R \citep{han2024valler} \textsuperscript{\dag}  & 1.58 & 2.32 & 0.397 & 0.363 & 3.18 & 3.97 & 0.395 & 0.365 \\
   MELLE \citep{meng2024autoregressive} \textsuperscript{\ddag} & \textbf{1.53} & \textbf{2.22} & 0.517 & 0.480 & 2.21 & \textbf{2.80} & 0.633 & 0.591\\

   \midrule                 
   \our & \textbf{1.53} & 2.27 & \textbf{0.539} & \textbf{0.513}  & \textbf{2.20} & 2.89 & \textbf{0.654} & \textbf{0.619} \\

\bottomrule
\end{tabular}

}

\end{table*}

\section{Results}

\subsection{Comparative Study}

This section presents a comprehensive comparison between our proposed approach and existing TTS systems, using predicted MOS scores to assess perceptual quality and objective metrics such as WER and similarity to evaluate linguistic accuracy and fidelity.

Table~\ref{tab:ramp_mos} summarizes the predicted MOS evaluations for various TTS systems on the LibriSpeech test-clean dataset, assessed under two distinct conditions: continuation and cross-sentence scenarios. In continuation tasks, MELLE emerges as the top-performing system with an MOS of 3.843, slightly exceeding our proposed method. Both systems exhibit near-human-level performance, closely approximating the ground truth MOS of 4.043. These results substantiate the effectiveness of autoregressive models based on continuous representation in capturing intra-sentence continuity. However, VALL-E demonstrates substantially inferior performance, revealing its limitations in preserving contextual coherence within utterances. When transitioning to the cross-sentence task, our proposed method achieves remarkable performance with an MOS of 4.157, outperforming both the MELLE baseline and the human ground truth. This superior performance underscores our architecture's advanced capacity in modeling long-range dependencies.

From the perspective of word error rate and similarity, we compare \our{} with several speech synthesis models, including ELLA-V, VALL-E R, RALL-E, CLAM-TTS, VALL-E, and MELLE in Table~\ref{tab:obj_1}. These models are strong baselines in the field of speech synthesis. These models represent a diverse range of approaches, encompassing both discrete representation-based and continuous representation-based methods, providing a valuable benchmark for evaluating our approach. Additionally, the performance of ground truth (GT) is included as a reference point to highlight the upper bounds of these metrics. Another GT, labeled as GT-mel, refers to the reconstruction from the mel-spectrogram with the vocoder.

VALL-E provides the benchmark performance of zero-shot TTS based on language model modeling. Models like ELLA-V and VALL-E R achieve relatively low WER scores but perform poorly in similarity metrics, highlighting a trade-off between transcription accuracy and speech similarity. CLAM-TTS struggles to balance the continuation and cross-sentence tasks. Among the models compared, MELLE stands out by achieving substantial improvements in both WER and similarity metrics across two settings. In particular, WER scores of MELLE are not only lower than most competing models but even exceed the ground truth levels, especially in continuation settings. This indicates an exceptional ability to capture linguistic accuracy. However, since WER has already reached near-optimal levels in MELLE, further improvement in similarity metrics becomes increasingly critical for advancing speech synthesis performance. Our proposed model, \our{}, maintains the strong WER performance demonstrated by MELLE, achieving similarly low WER scores in both continuation and cross-sentence tasks. Importantly, \our{} delivers significant improvements in similarity metrics. These results indicate that \our{} has the potential to advance the balance between transcription accuracy and speech similarity.
\begin{table*}


\caption{The ablation study from two perspectives: the prior distribution and the generation mechanism. The \cmark denotes methods used in our paper, while others refer to ablation methods. `Vanilla Prior' denotes a standard Gaussian distribution as the prior, `HFM' denotes holistic flow matching that generates complete mel-spectrogram frames through a unified flow-matching process, and `DFM' represents decoupled flow matching where separate flow-matching processes independently generate low- and high-frequency components without cross-band condition.}
  \label{tab:ablation}

  \centering 
    \resizebox{0.8\textwidth}{!}
  {
  \begin{tabular}{ccc cccc cccc }
  \toprule
 \multirow{2}{*}{\textbf{\makecell[c]{Prior \\Distribution}}}   & \multirow{2}{*}{\textbf{\makecell[c]{Generation\\Mechanism}}}  & \multicolumn{4}{c}{\textbf{Continuation}} & \multicolumn{4}{c}{\textbf{Cross-Sentence}} \\
 \cmidrule(r){3-6} \cmidrule(r){7-10}
  &   & WER-C  & WER-H  & SIM-r & SIM-o   & WER-C  & WER-H &  SIM-r & SIM-o   \\
  \midrule

\cmark    & \cmark  & \textbf{1.53} & \textbf{2.27} & \textbf{0.539} & \textbf{0.513}  & \textbf{2.20} & \textbf{2.89} & \textbf{0.654} & \textbf{0.619}    \\
 Vanilla Prior   & \cmark  & 1.72 & 2.53 & 0.502 & 0.466 &  2.72  & 3.56 & 0.627 & 0.580         \\
  \cmark   &  HFM &  1.78& 2.35& 0.484 & 0.451 & 2.82 & 3.81 & 0.579 & 0.536   \\ 
  \cmark   &  DFM  &  1.88&  2.74&   0.492&   0.451&  3.66 & 4.51  & 0.622  &  0.575  \\ 
\bottomrule 	 		 	 	 	 
\end{tabular}

}

\end{table*}
\begin{figure}[t]
  \includegraphics[width=\columnwidth]{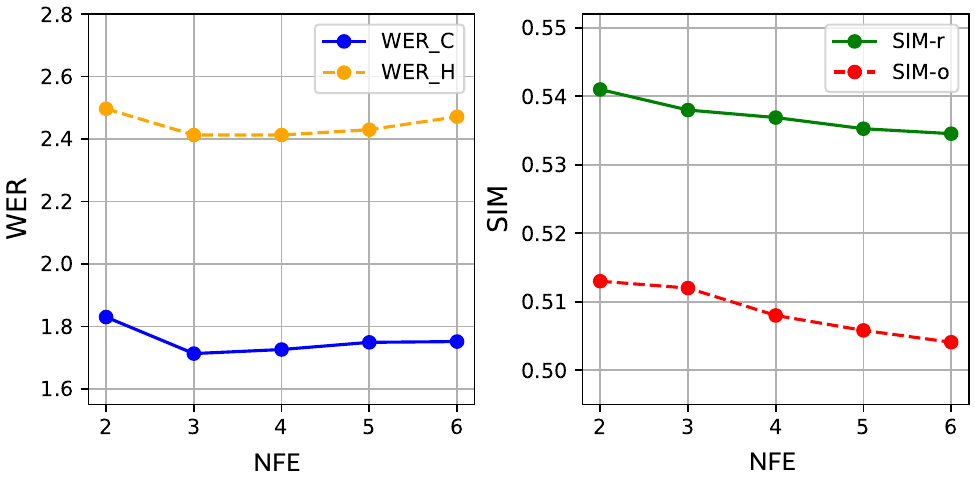}
  \caption{Continuation performance across different NFEs. Results are averaged over multiple runs and checkpoints.}
  \Description{Bar chart showing continuation performance at different numbers of function evaluations (NFEs). Each bar represents an average over multiple runs and model checkpoints. Performance varies with the number of NFEs.}
  \label{fig:stepeffect}
\end{figure}

\subsection{Ablation Study}

The results presented in Table~\ref{tab:ablation} explore the importance of the prior distribution and the generation mechanism in achieving optimal performance. 

In terms of the prior distribution, we compare two prior initialization approaches: (1) using the previous frame as the prior distribution versus (2) employing the Vanilla Prior (Gaussian initialization), which is the conventional baseline. The experimental results, as shown in the first two rows of Table~\ref{tab:ablation}, demonstrate that Vanilla Prior results in higher error rates and lower similarity scores, indicating a degradation in both accuracy and semantic coherence. Notably, the flow-matching process with our learned prior achieves optimal performance in 3 flow-matching steps, whereas the Vanilla Prior requires over 7 steps. This demonstrates that prior optimization significantly improves computational efficiency.

Regarding the generation mechanism, both ablated methods, Holistic Flow Matching (HFM) and Decoupled Flow Matching (DFM), underperform compared to our proposed approach. While DFM shows better similarity metrics than HFM through independent low-/high-frequency generation, it simultaneously increases WER. This reveals an inherent trade-off: DFM's decoupled architecture enhances local feature matching but lacks cross-band coordination, damaging semantic coherence. Our frequency-conditioned method overcomes this limitation by establishing dynamic spectral interactions, achieving an optimal balance between detail fidelity and structural consistency.

\subsection{Parameter Analysis} 

In this section, we conduct a detailed analysis of several critical parameters, including the number of function evaluations (NFE), classifier-free guidance scale, the scaling of the C2F-FM network, and prior variance.

\paragraph{Number of Function Evaluations} The impact of the number of function evaluations on TTS performance is shown in Figure~\ref{fig:stepeffect}. In our experiments, we employ the Euler method for numerical integration, where each step corresponds to one function evaluation. Initially, WER decreases as NFE increases, reflecting the improved mapping between prior distributions and acoustic outputs achieved by flow matching. This enhancement leads to better intelligibility and clarity in the generated speech. However, beyond a certain threshold, WER begins to rise, likely due to overfitting or the introduction of excessive distortion caused by additional iterations. In contrast, the similarity metric consistently declines as NFE increases. This suggests that excessive refinement may result in over-smoothing or the loss of fine-grained details, reducing perceptual similarity. These observations highlight a critical trade-off in the use of flow matching: while a moderate number of NFE improves clarity and reduces WER, excessive NFE can degrade both intelligibility and similarity. Therefore, careful tuning of NFE is essential to strike an optimal balance between clarity and similarity.

\begin{figure}[t]
  \includegraphics[width=\columnwidth]{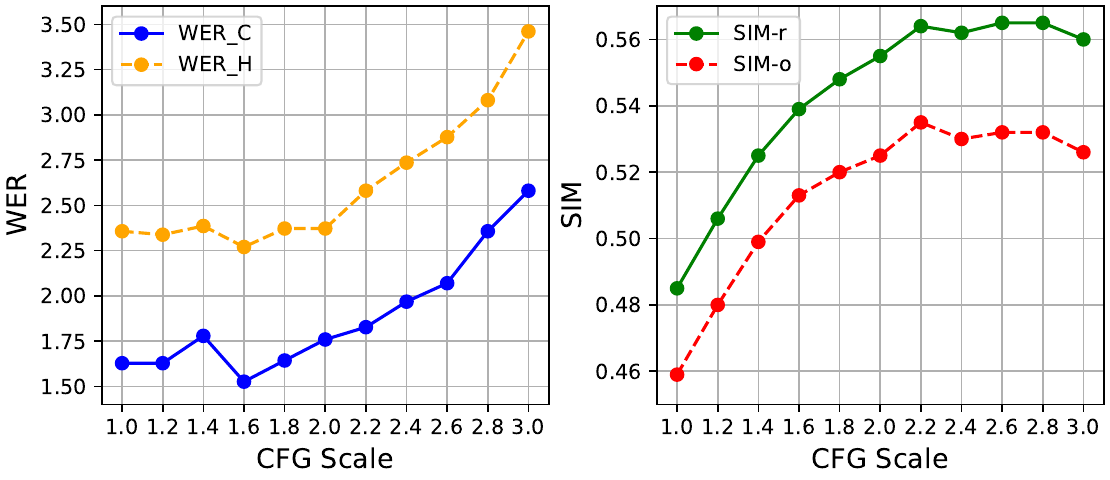}
  \caption{Continuation performance across CFG scales (1 to 3, step 0.2). CFG scale 1 means no CFG applied.}
  \Description{Line plot showing continuation performance as CFG scale increases from 1 to 3 in steps of 0.2. The performance generally improves with higher CFG scale, starting from scale 1, which represents no CFG applied.}
  \label{fig:scaleeffect}
\end{figure}

\begin{table*}

  \footnotesize

  \caption{Comparison of performance with the scale of the C2F-FM Network on \textit{continuation} and \textit{cross-sentence} zero-shot speech synthesis tasks. The first column indicates the number of residual blocks and the width of fully-connected layers in each block (e.g., 3×512 denotes 3 residual block with 512-dimensional layers). Total parameters (\#~Params) represent the sum of trainable parameters in both coarse and fine flow-matching networks.}
  \label{tab:FMnet}

  \centering 
    \resizebox{0.8\textwidth}{!}
  {
  \begin{tabular}{ cc cccc cccc}

    \toprule
 \multirow{2}{*}{\textbf{Scale}} &\multirow{2}{*}{\textbf{\# Params}}& \multicolumn{4}{c}{\textbf{Continuation}} & \multicolumn{4}{c}{\textbf{Cross-Sentence}} \\
 \cmidrule(r){3-6} \cmidrule(r){7-10}
     & & WER-C  & WER-H  & SIM-r & SIM-o   & WER-C  & WER-H &  SIM-r & SIM-o   \\
 \midrule
   3×512& 6M & 1.61& 2.28& 0.487 & 0.458  &2.45 & 3.15 &0.590  &    0.553    \\
   3×1024& 18M  & 1.53 & 2.27 & 0.539 & 0.513  & 2.20 & 2.89 & 0.654 & 0.619 \\
   6×1024& 34M  & 1.63 & 2.43 & 0.520 & 0.490  & 2.28 & 2.91 & 0.616 & 0.581 \\
   12×1024& 77M &1.55 &2.34  &0.500 &0.475  &2.30 &3.04 &0.611  & 0.579  \\		 	 		 	 					
\bottomrule
\end{tabular}

}

\end{table*}

\paragraph{CFG Scale} Figure~\ref{fig:scaleeffect} illustrates the impact of the CFG scale \(w\) on TTS performance. The experimental results reveal that employing an optimal CFG scale yields significant performance enhancements compared to the baseline without CFG, particularly in speech similarity. The WER trend shows a consistent decrease as the CFG scale increases from 1.0 to 1.6, suggesting better text-speech alignment and improved intelligibility. However, beyond the threshold of 1.6, the WER exhibits an upward trend. Regarding speech similarity, the metric reaches its optimal value at a CFG scale of 2.2, after which it deteriorates due to the over-constraining effect that compromises speech naturalness. These observations indicate that WER and similarity metrics demonstrate distinct response patterns to CFG scale variations, underscoring the importance of meticulous parameter tuning to achieve an optimal balance between speech intelligibility and naturalness in TTS systems.

\paragraph{Scaling of the C2F-FM Network} The results in Table \ref{tab:FMnet} highlight the trends of WER and similarity as the scale of the C2F-FM network increases. The smallest model demonstrates limited learning capacity, resulting in high WER and poor similarity scores, suggesting the limitations of model size on its capabilities. Further scaling to 3×1024 achieves a better balance, with both WER and similarity metrics showing notable improvements. However, when the model size is increased further to 6×1024 and 12×1024, the gains in WER become marginal, and similarity scores begin to degrade, indicating potential overfitting. This suggests that while larger models can capture more complex patterns, excessive scaling may harm generalization, particularly in maintaining high similarity. In conclusion, simply increasing the scale of the C2F-FM network is not sufficient to achieve optimal performance. Future work should focus on developing more efficient architectures and training strategies to improve both WER and similarity without overfitting, ensuring better generalization in zero-shot speech synthesis tasks.








\begin{table}
  \centering
  \caption{Effect of prior distribution variance (\(\sigma^2\)) on zero-shot speech synthesis tasks.}
  \resizebox{0.75\columnwidth}{!}{ 
  \begin{tabular}{ccccc}
     
    \multicolumn{5}{c}{\textbf{Continuation Task}} \\
    \toprule
    \(\sigma^2\) & WER-C & WER-H & SIM-r & SIM-o \\
    \midrule
    0.10 & 1.53 & 2.27 & 0.539 & 0.513 \\
    0.15 & 1.48 & 2.20 & 0.524 & 0.499 \\
    0.20 & 1.48 & 2.14 & 0.502 & 0.478 \\
    \bottomrule
    
    \addlinespace[0.5em]
    \multicolumn{5}{c}{\textbf{Cross-Sentence Task}} \\
   \toprule
    \(\sigma^2\) & WER-C & WER-H & SIM-r & SIM-o \\
    \midrule
    0.10 & 2.20 & 2.89 & 0.654 & 0.619 \\
    0.15 & 2.28 & 3.07 & 0.628 & 0.596 \\
    0.20 & 2.20 & 2.91 & 0.600 & 0.570 \\
    \bottomrule
  \end{tabular}}
 
  \label{tab:initD}
\end{table}
\paragraph{Prior Variance}
The experiments in Table~\ref{tab:initD} reveal that increasing the prior variance \(\sigma^2\) in Equation~\ref{eq:prior} introduces a systematic trade-off between synthesis accuracy and prosodic continuity. For continuation tasks, higher \(\sigma^2\) values yield improved WER, suggesting that moderate noise injection enhances the model’s adaptability to contextual variations. This improvement, however, coincides with reduced similarity scores, indicating that while the model becomes more robust to token transitions, excessive variance weakens temporal coherence between adjacent frames. In cross-sentence synthesis, \(\sigma^2=0.1\) achieves optimal performance. While WER values remain stable across \(\sigma^2\) settings and similarity metrics degrade sharply at higher \(\sigma^2\) (e.g., \(\sigma^2=0.2\)). This underscores that cross-sentence continuity relies on tighter prior conditioning (lower \(\sigma^2\)) to preserve acoustic relationships between sentence boundaries.

\section{Conclusion}
In this paper, we propose a novel autoregressive speech synthesis framework based on continuous representations, which overcomes the limitations of temporal consistency and model capacity in existing systems. By leveraging the sequential nature of language models and the temporal dynamics of speech signals, \our{} utilizes pervious tokens to assist in the flow-matching generation process. A coarse-to-fine flow-matching architecture is then developed, capturing both temporal and spectral correlations present in mel-spectrograms, allowing for precise modeling of each continuous token. Experimental results show that our model consistently outperforms several baseline systems across various evaluation metrics, producing clear and natural speech with significantly improved similarity.

\clearpage

\begin{acks}
This work has been supported by the National Key R\&D Program of China through grant 2022ZD0116307 and NSF China (Grant No.62271270).
\end{acks}

\bibliographystyle{ACM-Reference-Format}
\balance
\bibliography{sample-base}

\end{document}